\documentclass[letterpaper]{article} % DO NOT CHANGE THIS
\usepackage{aaai25}  % DO NOT CHANGE THIS
\usepackage{times}  % DO NOT CHANGE THIS
\usepackage{helvet}  % DO NOT CHANGE THIS
\usepackage{courier}  % DO NOT CHANGE THIS
\usepackage[hyphens]{url}  % DO NOT CHANGE THIS
\usepackage{graphicx} % DO NOT CHANGE THIS
\usepackage{natbib}  % DO NOT CHANGE THIS AND DO NOT ADD ANY OPTIONS TO IT
\usepackage{caption} % DO NOT CHANGE THIS AND DO NOT ADD ANY OPTIONS TO IT
\usepackage{multirow}
\usepackage{booktabs}
\usepackage{amsmath}
\usepackage{bbding}
\usepackage{color}
\usepackage[ruled,vlined,linesnumbered,noresetcount]{algorithm2e}
\SetKwRepeat{Do}{do}{while}
\SetKw{And}{and}
\SetKw{In}{in}

\title{MDD-5k: A New Diagnostic Conversation Dataset for Mental Disorders Synthesized via Neuro-Symbolic LLM Agents}
\author{
    Congchi Yin\textsuperscript{\rm 1,2}\thanks{Work done during an internship.},
    Feng Li\textsuperscript{\rm 1},
    Shu Zhang\textsuperscript{\rm 1},
    Zike Wang\textsuperscript{\rm 1}\footnotemark[1],
    Jun Shao\textsuperscript{\rm 1}\thanks{Corresponding authors.},
    Piji Li\textsuperscript{\rm 2},\\
    Jianhua Chen\textsuperscript{\rm 3,4,5,6}\footnotemark[2],
    Xun Jiang\textsuperscript{\rm 1,2}
}
\affiliations{
    \textsuperscript{\rm 1}Theta Health Inc.
    \textsuperscript{\rm 2}Chen Frontier Lab for AI and Mental Health, Tianqiao and Chrissy Chen Institute, Shanghai, China\\
    \textsuperscript{\rm 3}Shanghai Mental Health Center\\
    \textsuperscript{\rm 4}Shanghai Jiao Tong University School of Medicine\\
    \textsuperscript{\rm 5}Shanghai Clinical Research Center for Mental Health\\
    \textsuperscript{\rm 6}Shanghai Key Laboratory of Psychotic Disorders\\

    congchi.yin@gmail.com, \{jun.shao, xun.jiang\}@thetahealth.ai, jianhua.chen@smhc.org.cn
}

\begin{document}

\maketitle
% \renewcommand{\thefootnote}{\fnsymbol{footnote}}
% \footnotetext[2]{Work done during an internship.}
% \footnotetext[3]{Corresponding authors.}

\begin{abstract}
The clinical diagnosis of most mental disorders primarily relies on the conversations between psychiatrist and patient. The creation of such diagnostic conversation datasets is promising to boost the AI mental healthcare community. However, directly collecting the conversations in real diagnosis scenarios is near impossible due to stringent privacy and ethical considerations. 
To address this issue, we seek to synthesize diagnostic conversation by exploiting anonymized patient cases that are easier to access. 
Specifically, we design a neuro-symbolic multi-agent framework for synthesizing the diagnostic conversation of mental disorders with large language models. It takes patient case as input and is capable of generating multiple diverse conversations with one single patient case. The framework basically involves the interaction between a doctor agent and a patient agent, and generates conversations under symbolic control via a dynamic diagnosis tree. 
By applying the proposed framework, we develop the largest Chinese mental disorders diagnosis dataset MDD-5k. This dataset is built upon 1000 real, anonymized patient cases by cooperating with Shanghai Mental Health Center and comprises 5000 high-quality long conversations with diagnosis results and treatment opinions as labels. To the best of our knowledge, it's also the first labeled dataset for Chinese mental disorders diagnosis. Human evaluation demonstrates the proposed MDD-5k dataset successfully simulates human-like diagnostic process of mental disorders.
\begin{links}
\link{Code\&Dataset}https://github.com/lemonsis/MDD-5k.
\end{links}
\end{abstract}

\section{Introduction}
Mental health issues have garnered increasing attention in recent years. According to the statistics of World Health Organization (WHO), one in every eight people in the world lived with a mental disorder in 2019, and people living with anxiety and depressive disorders kept rising significantly because of the COVID-19 pandemic \citep{world2022mental}. 
With the recent progress of large language models (LLMs) \citep{ouyang2022training}, which emerges capabilities of human-like text generation, many researchers turn to building conversational AI system for mental healthcare. Current implementations can be divided into two categories, finetuning a small model (e.g. Llama2-7B \citep{touvron2023llama}) with physician-patient conversations \citep{EmoLLM,Yang_2024,liu2023chatcounselor} or building prompt-based physician-patient role-playing framework \citep{wang2024notechatdatasetsyntheticdoctorpatient,zhang2024cpsycoun} with state-of-the-art language models (e.g. ChatGPT \citep{ouyang2022training}). Regardless of the method employed, domain-specific mental health datasets play a fundamental and indispensable role. 

\begin{figure}
\centering
\includegraphics[width=\columnwidth]{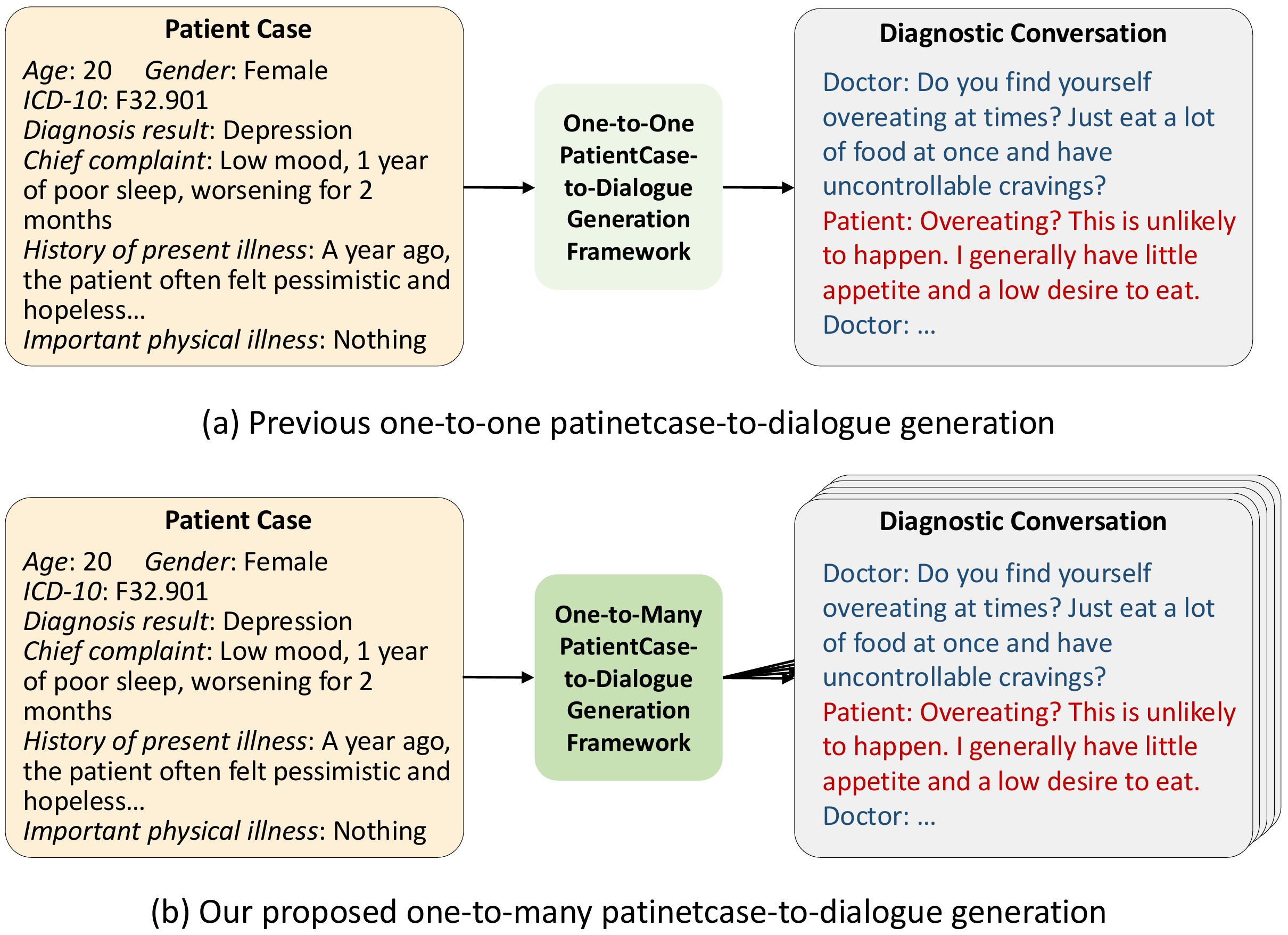}
\caption{Illustration of previous one-to-one patientcase-to-dialogue generation and our proposed one-to-many-patientcase-to-dialogue generation framework.}
\label{f1}

\end{figure}

We focus on diagnostic conversation dataset for mental disorders in this work. The clinical diagnosis of mental disorders differs from other diseases in that it primarily relies on the mental status examination of patients, which is reflected through conversations between psychiatrists and patients rather than physiological indices \citep{first2014structured}. Therefore, the collection of mental disorders diagnostic conversation is promising to facilitate a variety of downstream tasks in AI mental health research like auxiliary diagnosis chatbot, mental disorders classification, etc.
However, while many previous studies \citep{sun2021psyqachinesedatasetgenerating,chen-etal-2023-soulchat} focused on emotional support or psychological counseling data, few work shed light on diagnostic conversations of mental disorders. This can be attributed to two main factors. First, diagnostic conversations in real scenarios are extremely hard to acquire due to the privacy and ethical consideration. Second, synthesizing diagnostic conversations from scratch is also challenging. Unlike psychological counseling or empathetic dialogue, diagnosis follows standardized process and requires professional medical knowledge. Consequently, directly employing LLMs for data synthesis in this context often yields poor outcomes \citep{tu2024towards}.
$\text{D}^4$ \citep{yao-etal-2022-d4} made the first attempt by simulating diagnostic conversations with employed annotators. However, it only covers depressive disorder and entirely depends on human annotation. The generated content is also short and far from similar to diagnostic conversations in real scenarios.

We propose a neuro-symbolic multi-agent framework that takes patient cases as input to synthesize diagnostic conversations of mental disorders. The framework involves three types of large language model agents: a doctor agent, a patient agent, and a symbolic tool agent responsible for managing diagnostic topic shift. This framework features two major innovations:
(\romannumeral1) One-to-many patientcase-to-dialogue generation that maximizes the utilization of precious real patient cases. As shown in Figure \ref{f1}, unlike previous studies \citep{zhang2024cpsycoun,wang2024notechatdatasetsyntheticdoctorpatient} that generate one conversation with one patient case. Our proposed framework is capable of generating multiple diverse diagnostic conversations with one single patient case. Specifically, three methods ensure the diversity and correctness of diagnostic process. 
First, doctor agents with different diagnosis habits are designed and randomly selected for each conversation. 
Second, we use LLM with knowledge graph to generate multiple fictitious patient experiences given one patient case. The patient experiences serve as background information for patient agents during generation. Since the diagnosis of mental disorders mainly relies on symptoms rather than concrete events, integrating the fictitious patient experiences enhances the diversity of synthesized conversation while maintaining the accuracy of the diagnostic process.
Third, the sequence of diagnostic topics is randomly determined for each conversation.
(\romannumeral2) Another significant innovation lies in text generation under symbolic control via a dynamic diagnosis tree. This tree consists of a fixed symptom inquiry tree and a dynamic experience inquiry tree. Clinical diagnosis of mental disorders strictly follows standards from ICD-11 \citep{world2018icd} or DSM-5 \citep{american2013diagnostic}. To simulate this process, we design a fixed symptom inquiry tree based on Structured Clinical Interview for DSM-5 (SCID-5) \citep{first2014structured}, covering all the diagnostic topics for important symptoms inquiry. The experience inquiry tree is constructed by extracting possible topics from patient's response of past experiences. It's designed to establish deeper engagements with the patient. 

By applying the proposed framework, we release the largest Chinese \textbf{M}ental \textbf{D}isorder \textbf{D}iagnosis dataset MDD-5k. It's also the first labeled mental disorders diagnostic conversation dataset with diagnosis results from professional psychiatrists.
MDD-5k contains 5000 high-quality diagnostic conversations and is built upon 1000 anonymized, real patient cases from Shanghai Mental Health Center, covering over 25 different diseases. All the patient cases are cleaned and filtered in accordance with global standards to ensure the complete protection of private information. 

The contributions of this work can be summarized as:
\begin{itemize}
    \item We specially design a neuro-symbolic multi-agent framework for synthesizing diagnostic conversation of mental disorders, which features controllable and diverse one-to-many patientcase-to-dialogue generation.
    \item We propose MDD-5k which is the largest and first labeled Chinese mental disorders diagnosis dataset to the best of our knowledge.
    % \item The largest Chinese mental disorders diagnosis dataset MDD-5k is proposed, which contains 5000 high-quality long conversations with convincing diagnosis results as labels. To the best of our knowledge, it's also the first mental disorders diagnosis dataset with labels.
    \item Comprehensive human evaluation shows the proposed MDD-5k dataset outperforms several compared datasets in professionalism, communication skills, fluency, safety, and mirrors human-like diagnostic process.
\end{itemize}

\section{Related Work}
\subsection{Mental Health Dataset}
Corpora of physician-patient conversations focused on mental health are crucial for AI mental healthcare research, especially in the large language model era. We divide current mental health datasets into three categories based on the degree of required professional knowledge. \textit{Emotional support datasets} feature empathetic dialogue and comfort. ESconv dataset \citep{liu-etal-2021-towards} consists of 1300 conversations covering 10 topics. SoulChatCorpus \citep{chen-etal-2023-soulchat} contains over 2 million single-turn and multi-turn conversations generated by ChatGPT.
\textit{Psychological counseling datasets} typically contain more domain knowledge than common emotional support dataset. The Emotional First Aid Raw dataset \citep{EfaqaCorpusRaw:chatopera2024} is built by crawling from psychological counseling websites and communities. PsyQA \citep{sun2021psyqachinesedatasetgenerating} is a single-turn Chinese dataset annotated by human. SmileChat dataset \citep{qiu2024smilesingleturnmultiturninclusive} expands PsyQA to multi-turn through ChatGPT. CPsyCounD \citep{zhang2024cpsycoun} contains 3134 counseling conversations generated by the same number of psychological counseling reports.
\textit{Diagnosis datasets} aim to simulate diagnostic conversation of professional psychiatrists. $\text{D}^4$ \citep{yao-etal-2022-d4} is a Chinese depression diagnosis dataset built by human annotators and supervised by psychiatrists.
There are also some medical dialogue datasets, like MedDialog \citep{he2020meddialog}, MTS-Dialog \citep{ben-abacha-etal-2023-empirical}, ChatDoctor \citep{li2023chatdoctor}, which encompass a broader range of medical fields.

\subsection{Mental Disorders Conversation Simulation}
We mainly focus on the tuning-free prompting frameworks for mental disorders conversation simulation. \citet{chen2023llm} conducted a comprehensive analysis on the feasibility of utilizing LLM chatbots in diagnostic conversation. \citet{wang2024patientpsi} proposed to simulate patient agent that integrates cognitive modeling with LLM, and applied this patient agent in cognitive behavior therapy (CBT) training. \citet{wang2024notechatdatasetsyntheticdoctorpatient} built a planning and role-playing method to generate dialogue from clinical note, and proposed a dataset of synthetic patient-physician conversations. \citet{zhang2024cpsycoun} introduced Memo2Demo framework which converts counseling report to counseling note and then applies it to generate conversations. \citet{tu2024towards} designed AIME framework which uses a self-play based simulated environment with automated feedback for diagnostic conversation generation.

\section{Methodology}
The synthesis process of mental disorders diagnostic conversations is presented. 
% The neuro-symbolic multi-agent framework features one-to-many patientcase-to-dialogue generation, indicating multiple generated conversations are based on the same patient case. 
As shown in Figure \ref{f2}, the framework basically involves the interactions among a doctor agent, a patient agent, and a tool agent. All the agents are played by large language models (LLMs). The doctor agent controlled by a dynamic diagnosis tree leads the diagnostic topic shift of the whole conversation. The patient agent responds to the doctor agent based on the preprocessed patient case and fictitious patient experience generated by the tool agent. The tool agent is also responsible for several symbolic operations of the dynamic diagnosis tree.

\subsection{Patient Cases Preprocessing}
The quality of patient cases is vital to the diagnostic conversation synthesis. We cooperate with Shanghai Mental Health Center and obtain over 1000 real cases of patients with mental disorders. All these patient cases have undergone data masking to prevent the leakage of sensitive personal information. The data masking process follows the standards below: (\romannumeral1) Private information of patients (e.g. name, date of birth, date of examination, etc.) is removed from the patient case. (\romannumeral2) Patient age is rounded to the nearest ten. For example, the age of a 24-year-old patient is 20 on the preprocessed patient case. (\romannumeral3) All the concrete locations are replaced with vague or fake ones. The above preprocessing steps strictly follow the Chinese information security technology guide for health data security (GB/T 39725-2020).

After filtering repetitive or incomplete patient cases, the final version for diagnostic conversation simulation and dataset generation contains 1000 patient cases with age, gender, diagnosis and corresponding International Classification of Diseases (ICD-10) \citep{world2004international} code, chief complaint, history of present illness, important past medical history, family history, personal history, mental examination, and treatment. As shown in Figure \ref{f2}, the patient case is structurized as key-value pairs. 
%We statistically analyzed some characteristics of preprocessed patient cases and results are illustrated in Figure \ref{}. 

\subsection{Fictitious Patient Experience Generation}
We perform one-to-many patientcase-to-dialogue generation, which indicates one patient case will be applied to generate multiple diagnostic conversations.                                                          
One key factor contributes to the diversity of generated conversations is the patient experience. It specifically refers to the past experiences that directly or indirectly lead to the mental illness problem of patients in this paper.
The diagnosis of mental disorders differs from other illnesses in that it mainly depends on the conversations between psychiatrists and patients instead of physiological indices \citep{first2014structured}. Psychiatrists provide diagnosis result and treatment based on the acquired symptoms of patients during communicating. As a result, if the correspondence between symptoms and diagnosis can be assured, the correctness and quality of the synthesized diagnostic conversation is guaranteed and is not affected by detailed patient experience. In this sense, it's feasible to generate multiple patient experiences with one patient case for synthesizing multiple diagnostic conversations.

\begin{figure*}[ht]
\centering
\includegraphics[width=\linewidth]{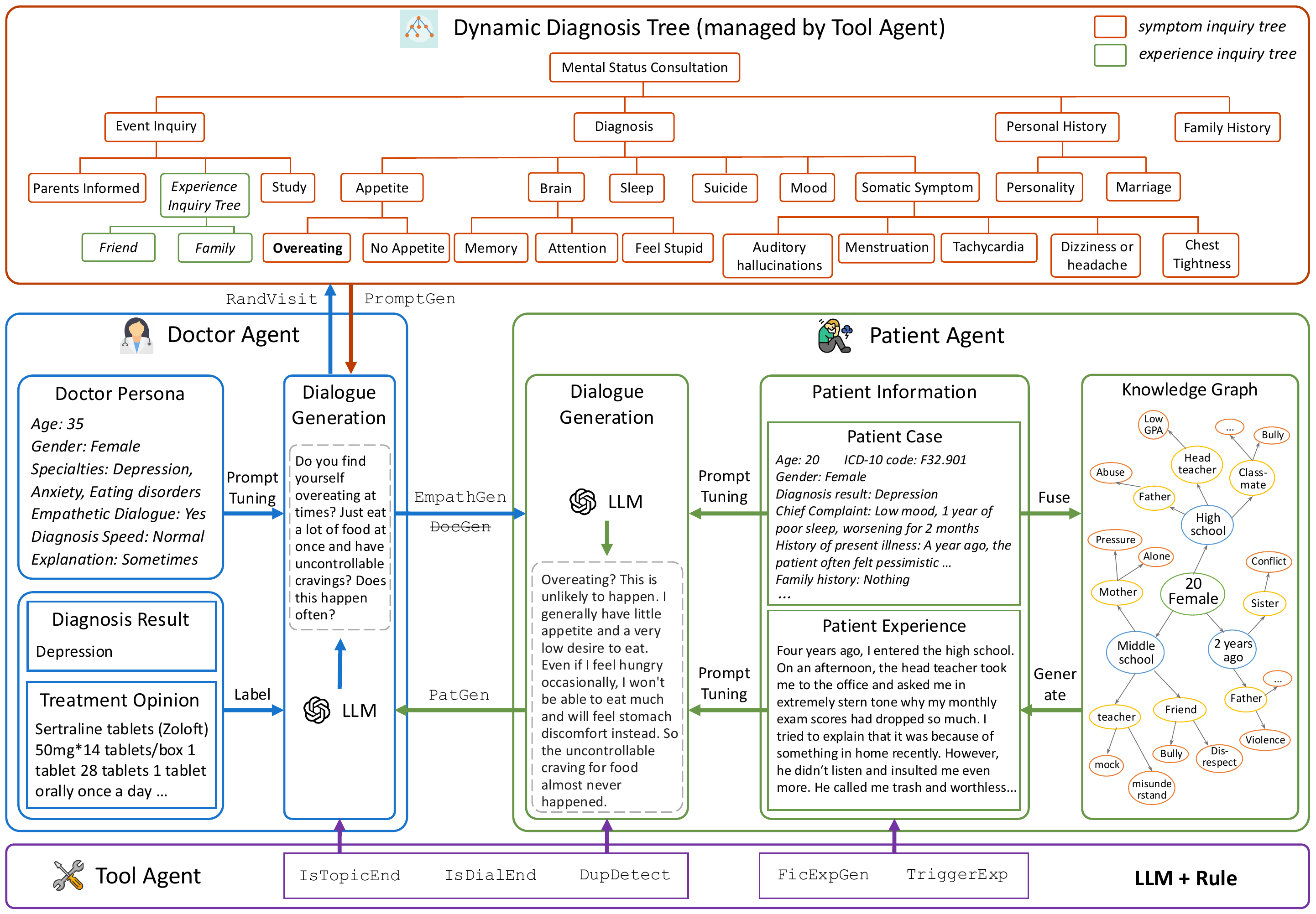}
\caption{The neuro-symbolic multi-agent LLM framework for synthesizing diagnostic conversation of mental disorders.}
\label{f2}

\end{figure*}

Large language model (LLM) is applied to generate fictitious patient experience. To avoid the counterfactual conflicts between fictitious patient experience and true patient case, gender, age, work and diagnosis (Dx) information from one patient case is extracted and serves as patient persona in the prompt for generating patient experience.
\begin{equation}
\label{1}
    \text{Persona} = \textit{Prompt}(\text{Gender}, \text{Age}, \text{Work}, \text{Dx})
\end{equation}
In Equation (\ref{1}), the function $\textit{Prompt}$ indicates concatenating keywords into proper prompt. Next, we build knowledge graphs containing time, people, and concrete event that might cause mental disorders according to different patient age and gender. The example in Figure \ref{f2} shows the predefined knowledge graph for 20-year-old female. The triplet $(\text{Time}, \text{People}, \text{Event})$ is randomly selected from the graph for fictitious experience generation.
\begin{equation}
    \text{FicExp} = \textit{Prompt}(\text{Time}, \text{People}, \text{Event})
\end{equation}
The final patient experience (FicExp) is generated through LLM by combining patient persona.
\begin{equation}
    \text{FicExp} = \textit{LLM}(\textit{Prompt}(\text{Persona}, \text{FicExp}))
\end{equation}

\subsection{Neuro-Symbolic Dynamic Diagnosis Tree}
To imitate conversations in real scenarios where the psychiatrist leads the entire diagnostic process, we design a neuro-symbolic \textit{dynamic diagnosis tree} to achieve diagnostic topic shift and controllable doctor response generation. The dynamic diagnosis tree consists of a \textit{symptom inquiry tree} and an \textit{experience inquiry tree}. As shown in the example of Figure \ref{f2}, the symptom inquiry tree is fixed and built according to the Structured Clinical Interview for DSM-5 (SCID-5) \citep{first2014structured} and guidance from professional psychiatrists. It aims to cover inquiries about all the relevant symptoms to arrive at the final diagnosis of a patient. Considering the gender and age differences, the symptom inquiry tree is specially designed for male and female, teenagers (people under 20), adults (people aged between 30 and 50) and elders (people over 60). The example in Figure \ref{f2} shows the symptom inquiry tree for female teenager.

The experience inquiry tree dynamically constructs itself based on patient's response regarding previous experiences and personal details. It is described as ``dynamic'' since each patient provides unique information about their background. A tool agent powered by LLM is responsible for parsing patient's response and creating corresponding topics which form the nodes of the experience inquiry tree. The parse process follows a depth-first manner. When the tool agent determines that the discussion around a specific topic is insufficient, it will keep parsing this topic to sub-topics until conversation around this topic is considered complete. Then it will move to the next parsed topic. The design of experience inquiry tree aims to establish deeper engagements with patients to facilitate diagnostic conversation.

The neuro-symbolic dynamic diagnosis tree is managed by the tool agent and offers \texttt{operations} for guiding both the doctor agent and the patient agent. Some operations are implemented by LLM and some are by rules. We first define five data types: Text, LNode, Tree, Graph, Bool. Text refers to natural language text. LNode stands for the leaf node of a dynamic diagnosis tree, indicating diagnostic topic in conversation. Tree refers to the hierarchical tree structure with a set of connected nodes. Although LNode can be viewed as a special Tree or Text, we treat it as a separate data type for clearer expression. Graph specifically refers to the knowledge graph for fictitious patient experience generation as explained before. Bool is a boolean variable which is either true or false. The operations for a doctor agent include:
\begin{itemize}
    \item \texttt{RandVisit}(\emph{$tr$}: Tree) $\rightarrow$ \emph{$ln$}: LNode. To improve the diversity of synthesized diagnostic conversation, we design the following leaf node visiting rules: (\romannumeral1) The parent nodes of these leaf nodes, representing high-level concept of diagnostic topic, are visited in a predefined order. (\romannumeral2) The leaf nodes under corresponding parent node, representing low-level specific diagnostic topic, are randomly visited. Visited leaf nodes will not be accessed again. \texttt{RandVisit} is responsible for implementing the above rules. It takes the whole dynamic diagnosis tree \emph{$tr$} as input and outputs one random leaf node \emph{$ln$}. This operation is implemented by rules.
    \item \texttt{IsTopicEnd}(\emph{$ln$}: LNode, \emph{$t$}: Text) $\rightarrow$ \emph{$b$}: Bool. It takes current diagnostic topic \emph{$ln$} and dialogue history \emph{$t$} around this topic as input, and decides whether conversation surrounding this topic should continue or end. This operation is implemented by LLM.
    \item \texttt{IsDialEnd}(\emph{$tr$}: Tree) $\rightarrow$ \emph{$b$}: Bool. If all the leaf nodes of the dynamic diagnosis tree \emph{$tr$} are visited, the operation will return true which marks the end of the diagnostic process. Else it will return false. This operation is implemented by rules.
    \item \texttt{ParseExp}(\emph{$t$}: Text) $\rightarrow$ \emph{$tr$}: Tree. The operation is responsible for building the dynamic experience inquiry tree. It takes patient responses \emph{$t$} containing experience information as input, and replaces the initial empty experience inquiry tree with a tree whose root node is \emph{$t$} and leaf nodes are possible topics related to \emph{$t$}. The output \emph{$tr$} is the updated dynamic diagnosis tree. This operation is implemented by rules and LLM.
    \item \texttt{DupDetect}(\emph{$t$}: Text, \emph{$tr^{(1)}$}: Tree) $\rightarrow$ \emph{$tr^{(2)}$}: Tree. As the diagnostic conversation progresses, some predefined topics may have already been discussed. \texttt{DupDetect} detects these duplicated topics in dialogue history \emph{$t$} and deletes them from the dynamic diagnosis tree to prevent repetitive conversation. It takes the original tree \emph{$tr^{(1)}$} as input and outputs an edited tree \emph{$tr^{(2)}$}. This operation is implemented by rules and LLM.
    \item \texttt{EmpathGen}(\emph{$ln$}: LNode, \emph{$t^{(1)}$}: Text) $\rightarrow$ \emph{$t^{(2)}$}: Text. In diagnostic conversation of mental disorders, the main goal of psychiatrist is to acquire symptoms from patients. Empathetic dialogue is not a must in this process. However, it sometimes helps in the diagnostic process and has been adopted by some doctors clinically. If the psychiatrist is accustomed to perform empathetic dialogue in daily consultations (this is reflected through the predefined doctor prompt which will be explained in the next subsection), \texttt{EmpathGen} will takes current diagnostic topic \emph{$ln$}, dialogue history \emph{$t^{(1)}$} as input and outputs comforting response \emph{$t^{(3)}$}. This operation is implemented by LLM.
    \item \texttt{PromptGen}(\emph{$ln$}: LNode) $\rightarrow$ \emph{$t$}: Text. It takes a leaf node \emph{$ln$} of the dynamic diagnosis tree as input and outputs proper prompt \emph{$t$} for instructing patient agent to respond around topic \emph{$ln$}. This operation is implemented by LLM.
\end{itemize}
The operations for a patient agent include:
\begin{itemize}
    \item \texttt{TriggerExp}(\emph{$t$}: Text) $\rightarrow$ \emph{$b$}: Bool. The operation decides whether it's time to trigger the \texttt{FicExpGen} operation or not, based on the doctor's question and dialogue history \emph{$t$}. This operation is implemented by rules and LLM.
    \item \texttt{FicExpGen}(\emph{$g$}: Graph, \emph{$t^{(1)}$}: Text) $\rightarrow$ \emph{$t^{(2)}$}: Text. The operation performs fictitious patient experience generation as detailedly described in the previous subsection. \emph{$g$} is the predefined knowledge graph. \emph{$t^{(1)}$} is the patient case and \emph{$t^{(2)}$} is the integrated patient information with real patient case and fictitious experience. This operation is implemented by LLM.
\end{itemize} 
The usage of these operations during agents interaction will be introduced in the next subsection.

\subsection{Conversation Synthesis with Agents}
The diagnostic conversation is synthesized by a doctor agent and a patient agent through role-playing LLMs. The doctor agent is under the guidance of the dynamic diagnosis tree. Initially, the dynamic diagnosis tree checks whether the current diagnostic topic should end. If affirmative, the doctor agent will turn to the next topic and check whether the future topics have been included in previous conversations. If not, the doctor agent will keep communicating with the patient around current topic. Then, if the patient talks about personal experience, the doctor agent will build the experience inquiry tree based on the patient response.

\begin{table}
\centering
\resizebox{\columnwidth}{!}
{
\begin{tabular}{ccccc}
\toprule
Dialogues & $\text{D}^4$ & CPsyCounD & Role-playing & MDD-5k$^*$\\
\midrule
Total num & 1339 & 3134& 100 & \textbf{5000}\\
Category & \textbf{diagnosis} &consultation&\textbf{diagnosis}&\textbf{diagnosis}\\
Illness &depression & / & / & \textbf{over 25} \\
Avg. turns &21.6&8.7&12.0 & \textbf{26.8}\\
Avg. words \#dial & 776 & 622.3& 1715.0 & \textbf{6906.8} \\
Avg. words \#doc &20.4 &49.7&88.1&\textbf{91.1} \\
Avg. words \#pat &14.9&30.4&47.6& \textbf{162.8}\\
Labels &\XSolidBrush &\XSolidBrush& / & \Checkmark \\
\bottomrule
\end{tabular}
}
\caption{Statistics of different datasets. Avg. words \#dial measures the average Chinese characters per dialogue. Avg. words \#doc and Avg. words \#pat measure the average Chinese characters per doctor's response and patient's response.}
\label{stat}

\end{table}

The doctor agent generates responses through operation \texttt{DocGen}(\emph{$ln$}: LNode, \emph{$t^{(1)}$}: Text) $\rightarrow$ \emph{$t^{(2)}$}: Text, which takes current diagnostic topic \emph{$ln$} and dialogue history \emph{$t^{(1)}$} as input and outputs response \emph{$t^{(2)}$}. To further improve the diversity of generated conversations, we design different diagnosis habits for the doctor agent. The diagnosis habits contain age, gender, specialties, empathetic dialogue, diagnosis speed, explanation, and serve as persona prompt for the doctor agent. An example is shown in Figure \ref{f2}. Specifically, the factors of empathetic dialogue and diagnosis speed exert huger effect on the doctor's response. If the doctor agent is accustomed to communicate empathetically, the \texttt{EmpathGen} operation introduced before will replace the \texttt{DocGen} operation for generation. If the diagnosis speed is set as fast, the doctor agent will speed up the diagnosis process, which leads to shorter conversations.

\begin{figure*}
\centering
\includegraphics[width=\linewidth]{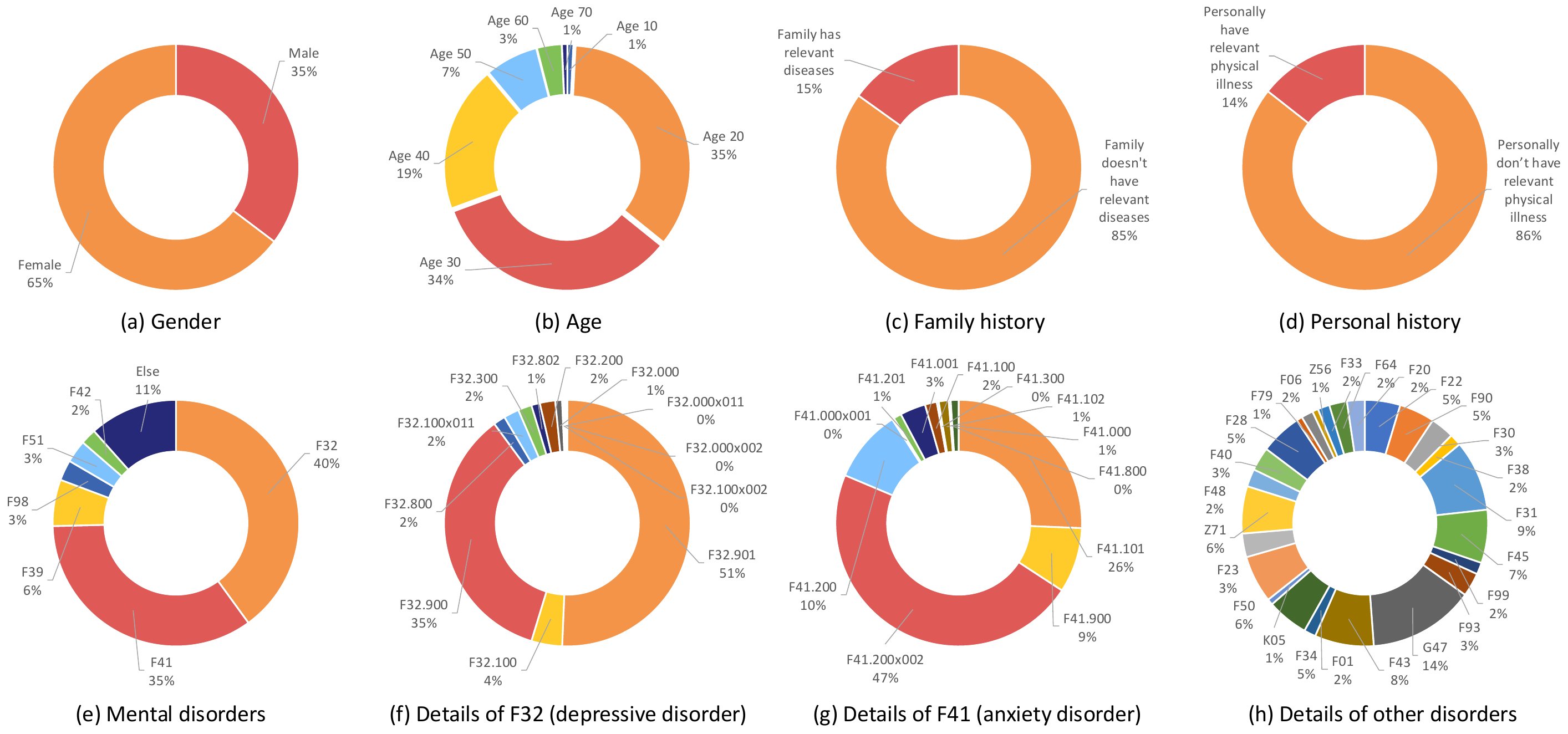}
\caption{Patient information of the MDD-5k dataset.}
\label{f3}

\end{figure*}

As to the patient agent, since the doctor agent leads the diagnosis process, the patient agent is designed to passively respond to the doctor based on known knowledge, including the patient case and generated fictitious experience. The operation \texttt{PatGen}(\emph{$ln$}: LNode, \emph{$t^{(1)}$}: Text, \emph{$t^{(2)}$}: Text) $\rightarrow$ \emph{$t^{(3)}$}: Text is responsible for patient response generation, which takes current diagnostic topic \emph{$ln$}, dialogue history \emph{$t^{(1)}$}, and patient case information \emph{$t^{(2)}$} as input and outputs proper response \emph{$t^{(3)}$}. If the patient agent determines to respond with personal experience under the control of dynamic diagnosis tree, fictitious patient experience will be fused into the patient case as input \emph{$t^{(2)}$}.
The whole process of the multi-agent framework for diagnostic conversation simulation of mental disorders is detailedly shown through pseudo-code in the Appendix.

\section{Experiment Setup}
\subsection{Implemention Details}
The MDD-5k dataset is generated through the neuro-symbolic multi-agent framework with 1000 real patient cases. 5 different fictitious patient experiences are generated by \texttt{gpt-4o} based on 1 patient case, which leads to a total of 5000 experiences corresponding to the 5000 conversations in the dataset. We also create 5 doctors with different diagnosis habits, and 1 doctor will be randomly picked for generating each conversation in the dataset. Since the patient cases are still under the ethical review, we randomly select 20 available patient cases and generate 100 conversations with \texttt{gpt-4o} for evaluation. The other 4900 conversations are currently generated by \texttt{Qwen2-72B-Instruct} \citep{qwen2} deployed on NVIDIA A100-80G GPUs locally.

\subsection{Compared Datasets}
As the evaluators' native language is Chinese, only Chinese datasets are considered to ensure the quality of human evaluation. Three datasets are selected as compared baselines. 
\begin{itemize}
    \item $\textbf{D}^4$ \citep{yao-etal-2022-d4} is a Chinese dialogue dataset for depression diagnosis, which is conducted by collecting conversations between professional psychiatrists.
    \item \textbf{CPsyCounD} \citep{zhang2024cpsycoun} is a synthetic consultation dataset covering nine representative topics (e.g. marriage, education) and seven classic schools of psychological counseling (e.g. cognitive behavioral therapy).
    \item \textbf{Direct Role-playing}: To test the effectiveness of our designed multi-agent diagnostic conversation simulation framework, we directly apply role-playing LLMs (\texttt{gpt-4o}) and generate 100 conversations with the same patient cases and prompts as MDD-5k for evaluation.
\end{itemize}
We haven't found any available open-source mental disorders diagnosis datasets besides $\text{D}^4$, so the consultation dataset CPsyCounD is chosen as baseline, which can also demonstrate the differences between psychological counseling and diagnostic conversation.
Statistics of these datasets are shown in Table \ref{stat}. We randomly select 100 samples from each dataset for evaluation.

\begin{table*}
\centering
\resizebox{\textwidth}{!}
{
\begin{tabular}{cccccccc}
\toprule
\textbf{Dataset} & \textbf{Professionalism} & \textbf{Communication (\romannumeral1)} & \textbf{Communication (\romannumeral2)} & \textbf{Fluency (\romannumeral1)} & \textbf{Fluency (\romannumeral2)}& \textbf{Similarity} & \textbf{Safety}\\
\midrule
$\text{D}^4$ & 6.6 & 7.9 & 7.8 & \textbf{8.6} & \textbf{8.2} & 7.2 & \textbf{0}\\
CPsyCounD & 5.2 & 5.4 & 5.6 & 8.4 & 8.0 & 4.4 & \textbf{0}\\
Role-playing & 6.8 & 6.6 & 7.2 & 6.9 & 5.5 & 6.4 & \textbf{0} \\
MDD-5k & \textbf{8.6} & \textbf{8.3} &\textbf{8.4} &  \textbf{8.6}& 7.6 &\textbf{8.8}& \textbf{0}\\
\bottomrule
\end{tabular}
}
\caption{Human evaluation of different datasets.}
\label{eva}

\end{table*}

\subsection{Evaluation Metrics}
Human evaluation is conducted to fairly assess the quality of different datasets. Specifically, we design seven major metrics encompassing five different perspectives: \textbf{professionalism}, \textbf{communication}, \textbf{fluency}, \textbf{similarity} and \textbf{safety}. 
\textbf{Professionalism} measures if the psychiatrist can effectively collect all the required patient symptoms for diagnosis.
\textbf{Communication} measures the psychiatrist's communication skills and the patient's response, including (\romannumeral1) Can the psychiatrist proactively ask patient for gathering key information and establish effective communication with the patients so that they are willing to share more information (e.g. daily life, past experiences) related to the mental illness problems? (\romannumeral2) Can the patient engage in the diagnostic process and tell related information? 
\textbf{Fluency} contains two aspects of criterion: (\romannumeral1) Are the generated conversations fluent in terms of both sentence and topic flow? (\romannumeral2) Is there any repetitive content or topic in the conversation?
\textbf{Similarity} measures how similar is the synthesized conversation to real scenarios.
The evaluators are guided to score from 1 to 10. A higher score indicates better performance.
\textbf{Safety} measures the leakage of private information (e.g. address). 0 means safe generation while 1 indicates privacy leakage.

Five annotators are employed for the human evaluation. Three of the them are professional psychiatrists who have years of clinical experience. The other two annotators are experienced in mental health data processing. The evaluation is conducted in a blind manner.

\section{Results and Evaluation}
\subsection{Statistical Analysis of MDD-5k}
The Figure \ref{f3} shows detailed patient information of the MDD-5k dataset. 65\% of the patients are female and 35\% are male. About 90\% of the patients are between 20 and 40 years old. 15\% of the patients report a family history of mental disorders and 14\% of the patients have relevant physical illness. Patients suffer from depressive disorder (F32) and anxiety disorders (F41) makes up of 75\% conversations of the dataset. Specifically, 51\% of the patients in depressive disorder are diagnosed with depressive state (F32.901), and 35\% of the patients are diagnosed with depressive episode (F32.900). 47\% of the anxiety disorder patients are diagnosed with anxiety and depression state (F41.200x002) and 26\% are diagnosed with anxiety state (F41.101). We also show details of other disorders which accounts for 11\% of the whole dataset in Figure \ref{f3}(h).
All the disease code follows standards in the second version of Chinese clinical classification of disease and codes. If a patient is diagnosed with multiple diseases, these diseases are counted separately.

The statistical details of MDD-5k and other compared datasets are presented in Table \ref{stat}. The MDD-5k dataset contains diagnostic conversations covering over 25 mental health illnesses. It includes 5000 dialogues, each comprising an averaged of 6906.8 Chinese characters which is almost ten times to compared datasets. The average dialogue turns are 26.8, slightly longer than the 21.6 turns of $\text{D}^4$. MDD-5k is also a labeled dataset with diagnosis result and treatment opinion from professional psychiatrist as label for each conversation, while $\text{D}^4$ only contains diagnosis results. 
Compared to the direct role-playing method without applying the multi-agent framework, the generated doctor response is about the same length. But the dialogue turns and patient response of MDD-5k are significantly longer, highlighting the effectiveness of our proposed framework in diagnostic conversation simulation. In the case study presented in Appendix, we show three complete samples of conversation with corresponding doctor persona, patient case, fictitious patient experience and the dynamic diagnosis tree.

\subsection{Human Evaluation}
The results of human evaluation are presented in Table \ref{eva}. MDD-5k exhibits superior performance across six major metrics. The evaluation scores on professionalism and similarity are significantly higher than other datasets, suggesting that our synthesized diagnostic conversations can mirror real scenarios of diagnosis to some extent. The communication quality of both doctor and patient is also impressive. Despite these strengths, MDD-5k does include some repetitive content, occasionally leading to less fluent conversations.
The $\text{D}^4$ dataset ranks second. It achieves relatively high score on communication and fluency evaluation. The biggest problem of $\text{D}^4$ is that its conversations are too brief and only include symptom inquiries and short responses.

An ablation study is conducted. The performance of the direct role-playing method is significantly worse compared to the neuro-symbolic multi-agent framework, particularly in terms of fluency and communication skills. This finding confirms that directly applying large language models for diagnostic conversation generation will lead to poor outcomes.
The evaluation also shows the distinct differences between diagnostic conversation and psychological counseling. The evaluation scores of CPsyCounD are notably low, especially for the professionalism and similarity metric. Psychological counseling prioritizes comfort and healing with different therapies, while diagnostic conversation focuses on acquiring symptoms to arrive at a final diagnosis result.

\section{Conclusion and Future Work}
We design a neuro-symbolic multi-agent framework for synthesizing diagnostic conversation of mental disorders, and apply it for building the first and largest open-source Chinese mental disorders diagnosis dataset with diagnosis results and treatment opinions as labels. The framework features controllable one-to-many patientcase-to-dialogue generation. Conversation between a doctor agent and a patient agent is guided by a dynamic diagnosis tree. We also employ several techniques to improve the diversity of generated conversations. Human evaluation shows the quality of the proposed MDD-5k dataset exceeds compared datasets on seven indicators. The MDD-5k dataset is believed to contribute to a wide range of downstream tasks like mental disorders classification, mental disorders diagnosis assistant training, etc.

The primary limitations of this work lie in three points: (\romannumeral1) The discrepancy between synthesized conversations and actual medical diagnostics remains a significant challenge. Large language models often struggle to interpret the full meaning of patient responses when they encapsulate diverse information aspects, consequently leading to redundant symptom inquiries. We are exploring various prompt strategies to mitigate this issue.
(\romannumeral2) We mainly design dynamic diagnosis tree for depression (F32), anxiety (F41), sleep disorders (F51), childhood emotional disorder (F98), and unspecified mood disorder (F39), which covers over 85\% conversations of MDD-5k. Nevertheless, some mental health conditions (e.g. obsessive-compulsive disorder (F42)) remain inadequately addressed, resulting in sub-optimal synthesized diagnostic conversation. Efforts are underway to expand our synthesis frameworks by designing more diagnosis trees to encompass a broader spectrum of mental disorders. 
(\romannumeral3) Only Chinese version of the MDD-5k dataset is proposed. We plan to translate it into English in the future.

\section*{Ethical Statement}
The collection of patient cases was conducted at the Shanghai Mental Health Center. All patients were informed that their information would be collected and used exclusively for research purposes. As detailed in the Patient Cases Preprocessing section, all data masking procedures strictly follow the Chinese information security technology guidelines for health data security (GB/T 39725-2020).
Currently, the patient cases and the synthesized diagnostic conversation dataset, MDD-5k, are undergoing an ethics review. We plan to release the MDD-5k dataset for research purposes only after the ethics review finishes. To prevent any potential privacy data leakage, all experiments are conducted on servers located within the Shanghai Mental Health Center.

\section*{Acknowledgments}
We are grateful for the GPU resources and accessible LLM API keys provided by Shanda Group. Chen Frontier Lab for AI and Mental Health, Tianqiao and Chrissy Chen Institute leads the project of collecting mental disorders patient cases by cooperating with Shanghai Mental Health Center.
We also appreciate the support and discussion from psychiatrists in Shanghai Mental Health Center.

\bibliography{aaai25}

\appendix
\section{Pseudo-code of the Proposed Framework}
We detailedly explain the multi-agent framework for diagnostic conversation simulation through pseudo-code. As illustrated in Algorithm \ref{al1}, the doctor agent, patient agent and tool agent are initialized with large language models. The whole process is guided by the tool agent.
When the diagnostic conversation finishes, the diagnosis result and treatment opinion from patient case will be appended to the dialogue history as label.

\begin{algorithm*}
    \caption{The neuro-symbolic multi-agent framework for synthesizing diagnostic conversation of mental disorders}
    \label{al1}
    \textbf{Initialize:} \textcolor{red}{DocAgent}, \textcolor{blue}{PatAgent}, \textcolor{cyan}{ToolAgent} with LLM. \emph{ddt}: Tree with predefined dynamic diagnosis tree. \emph{exp\_kg}: Graph with predefined knowledge graph. \emph{cur\_topic}: LNode with None. \emph{dial\_hist}, \emph{topic\_hist}: Text with empty list. \emph{pat\_info}: Text with patient case. \emph{treatment}: Text with treatment from patient case\;
    \While{\textbf{\emph{not}} \emph{\textcolor{cyan}{ToolAgent}}.\emph{\texttt{IsDialEnd}}\emph{(}ddt\emph{)}}{
        \uIf{cur\_topic is None}{
            \emph{cur\_topic} $\leftarrow$ \textcolor{cyan}{ToolAgent}.\texttt{RandVisit}(\emph{ddt})\;
            \emph{doc\_resp} $\leftarrow$ \textcolor{red}{DocAgent}.\texttt{DocGen}(\textcolor{cyan}{ToolAgent}.\texttt{PromptGen}(\emph{cur\_topic}), \emph{dial\_hist})\;
            \emph{dial\_hist}.append(\emph{doc\_resp}), \emph{topic\_hist}.append(\emph{doc\_resp})\;
            \uIf{\emph{\textcolor{cyan}{ToolAgent}}.\emph{\texttt{TriggerExp}}({dial\_hist})}{
                \emph{pat\_info} $\leftarrow$ \textcolor{cyan}{ToolAgent}.\texttt{FicExpGen}(\emph{exp\_kg}, \emph{pat\_info}) \;
            }
            \emph{pat\_resp} $\leftarrow$ \textcolor{blue}{PatAgent}.\texttt{PatGen}(\textcolor{cyan}{ToolAgent}.\texttt{PromptGen}(\emph{cur\_topic}), \emph{dial\_hist}, \emph{pat\_info})\;
            \emph{dial\_hist}.append(\emph{pat\_resp}), \emph{topic\_hist}.append(\emph{pat\_resp})\;
        }
        \uElse{
            \uIf{\emph{\textcolor{cyan}{ToolAgent}}.\emph{\texttt{IsTopicEnd}}\emph{(}cur\_topic, topic\_hist\emph{)}}{
                \emph{ddt} $\leftarrow$ \textcolor{cyan}{ToolAgent}.\texttt{DupDect}(\emph{topic\_hist}, \emph{ddt})\;
                \emph{topic\_hist} $\leftarrow$ [], \emph{cur\_topic} $\leftarrow$ \textcolor{cyan}{ToolAgent}.\texttt{RandVisit}(\emph{ddt})\;
                \uIf{cur\_topic = experience\_inquiry\_tree}{\emph{ddt} $\leftarrow$ \textcolor{cyan}{ToolAgent}.\texttt{ParseExp}(\emph{dial\_hist}),
                \emph{cur\_topic} $\leftarrow$ \textcolor{cyan}{ToolAgent}.\texttt{RandVisit}(\emph{ddt})\;
                \emph{doc\_resp} $\leftarrow$ \textcolor{red}{DocAgent}.\texttt{DocGen}(\textcolor{cyan}{ToolAgent}.\texttt{PromptGen}(\emph{cur\_topic}), \emph{dial\_hist})\;
                \emph{dial\_hist}.append(\emph{doc\_resp}), \emph{topic\_hist}.append(\emph{doc\_resp})\;
                \uIf{\emph{\textcolor{cyan}{ToolAgent}}.\emph{\texttt{TriggerExp}}({dial\_hist})}{
                \emph{pat\_info} $\leftarrow$ \textcolor{cyan}{ToolAgent}.\texttt{FicExpGen}(\emph{exp\_kg}, \emph{pat\_info}) \;
            }
            \emph{pat\_resp} $\leftarrow$ \textcolor{blue}{PatAgent}.\texttt{PatGen}(\textcolor{cyan}{ToolAgent}.\texttt{PromptGen}(\emph{cur\_topic}), \emph{dial\_hist}, \emph{pat\_info})\;
            \emph{dial\_hist}.append(\emph{pat\_resp}), \emph{topic\_hist}.append(\emph{pat\_resp})\;}
            \uElse{\emph{doc\_resp} $\leftarrow$ \textcolor{red}{DocAgent}.\texttt{DocGen}(\textcolor{cyan}{ToolAgent}.\texttt{PromptGen}(\emph{cur\_topic}), \emph{dial\_hist})\;\emph{dial\_hist}.append(\emph{doc\_resp}), \emph{topic\_hist}.append(\emph{doc\_resp})\;
                \uIf{\emph{\textcolor{cyan}{ToolAgent}}.\emph{\texttt{TriggerExp}}({dial\_hist})}{
                \emph{pat\_info} $\leftarrow$ \textcolor{cyan}{ToolAgent}.\texttt{FicExpGen}(\emph{exp\_kg}, \emph{pat\_info}) \;
            }\emph{pat\_resp} $\leftarrow$ \textcolor{blue}{PatAgent}.\texttt{PatGen}(\textcolor{cyan}{ToolAgent}.\texttt{PromptGen}(\emph{cur\_topic}), \emph{dial\_hist}, \emph{pat\_info})\;
            \emph{dial\_hist}.append(\emph{pat\_resp}), \emph{topic\_hist}.append(\emph{pat\_resp})\;}
            }
        \uElse{\emph{doc\_resp} $\leftarrow$ \textcolor{red}{DocAgent}.\texttt{DocGen}(\textcolor{cyan}{ToolAgent}.\texttt{PromptGen}(\emph{cur\_topic}), \emph{dial\_hist})\;
            \emph{dial\_hist}.append(\emph{doc\_resp}), \emph{topic\_hist}.append(\emph{doc\_resp})\;
            \uIf{\emph{\textcolor{cyan}{ToolAgent}}.\emph{\texttt{TriggerExp}}({dial\_hist})}{
                \emph{pat\_info} $\leftarrow$ \textcolor{cyan}{ToolAgent}.\texttt{FicExpGen}(\emph{exp\_kg}, \emph{pat\_info}) \;
            }
            \emph{pat\_resp} $\leftarrow$ \textcolor{blue}{PatAgent}.\texttt{PatGen}(\textcolor{cyan}{ToolAgent}.\texttt{PromptGen}(\emph{cur\_topic}), \emph{dial\_hist}, \emph{pat\_info})\;
            \emph{dial\_hist}.append(\emph{pat\_resp}), \emph{topic\_hist}.append(\emph{pat\_resp})\;}
        }
    }
    \emph{dial\_hist}.append(\emph{treatment})\;
    \Return \emph{dial\_hist}\;
\end{algorithm*}

% \section{Case Study}
% We present three cases of the synthesized diagnostic conversations in MDD-5k, as well as all the material for carrying out the conversation simulation, including patient case, doctor prompt, diagnostic topic shift, and generated fictitious patient experience. Diagnostic conversation 1 and 2 are generated with the same patient case 1 to highlight our one-to-many patientcase-to-dialogue framework is capable of generating diverse conversation. Dialogue conversation 3 is generate with another patient case 2.
% The red words are diagnosis result, namely the label for each conversation.
% Both Chinese and English samples are provided.
% As to the Chinese version,
% Figure \ref{apd1}, Figure \ref{apd5} and Figure \ref{apd8} show the patient cases and other related material. Figure \ref{apd2} to Figure \ref{apd4} show diagnostic conversation 1. Figure \ref{apd6} and Figure \ref{apd7} show diagnostic conversation 2. Figure \ref{apd9} to Figure \ref{apd11} show diagnostic conversation 3.
% As to the English version, Figure \ref{apd12} and Figure \ref{apd13}, Figure \ref{apd19} and Figure \ref{apd20}, Figure \ref{apd25} and Figure \ref{apd26} show the patient cases and other related material. Figure \ref{apd14} to Figure \ref{apd18} show diagnostic conversation 1. Figure \ref{apd21} to Figure \ref{apd24} show diagnostic conversation 2. Figure \ref{apd27} to Figure \ref{apd31} show diagnostic conversation 3.

\section{Feedback from Psychiatrists}
During human evaluation, psychiatrists also provide us feedback for the synthesized diagnostic conversation. The discussion between professional psychiatrists and the authors is listed.
\begin{itemize}
    \item The patient response is too ``sound and perfect''. In real diagnostic conversation, many mental disorders patients struggle to respond to the psychiatrists questions. For this issue, we are not sure whether a perfect patient answer or a poor but similar to real situation answer will benefit from the perspective of natural language processing.
    \item The proportion of patients suffer from severe depressive disorder is too small. The number of pediatric patients is also small.
    \item The synthesized dialogue lacks the presence of a companion role (e.g. parent), which is quite common in real scenarios. This actually inspires us to design a multi-party dialogue system for dealing with diagnostic conversation simulation in the future.
    \item Response from the doctor agent lacks proper explanation of symptoms to the patient. This is attribute to the poor mental health domain knowledge of large language models. Retrieve augmented generation (RAG) can be applied to alleviate this problem in the future.
    \item The diagnostic process is not detailed. For example, the doctor agent merely ask if the patient can sleep well in the evening when consulting symptoms related to sleep. More details like insomnia onset time, frequency, aggravation or relief should be inquired.
    However, patient case contains rather limited information. So designing such detailed questions might lead to fictional reply from large language models.
\end{itemize}
All the feedback is of great significance for future work in this domain.

\end{document}